\pdfoutput=1

\documentclass[11pt]{article}

\usepackage[]{emnlp2021}

\usepackage{times}
\usepackage{latexsym}

\usepackage{times}
\usepackage{latexsym}
\usepackage{caption}
\usepackage{subcaption}
\usepackage[T1]{fontenc}

\usepackage[utf8]{inputenc}

\usepackage{microtype}
\usepackage{appendix}
\usepackage{amsmath}
\usepackage{amssymb}
\usepackage{array,multirow,graphicx}
\usepackage{floatrow}
\usepackage[T1]{fontenc}
\usepackage{color}
\usepackage{colortbl}
\usepackage{soul}
\colorlet{soulred}{orange!30}
\sethlcolor{soulred}
\usepackage{microtype}

\author{George Chrysostomou \quad Nikolaos Aletras\\
  Department of Computer Science, University of Sheffield \\
  United Kingdom \\
  \texttt{\{gchrysostomou1, n.aletras\}@sheffield.ac.uk} \\}

\title{Enjoy the Salience: Towards Better Transformer-based Faithful Explanations with Word Salience}

\begin{document}
\maketitle

\begin{abstract}
     Pretrained transformer-based models such as BERT have demonstrated state-of-the-art predictive performance when adapted into a range of natural language processing tasks. An open problem is how to improve the faithfulness of explanations (rationales) for the predictions of these models. In this paper, we hypothesize that salient information extracted a priori from the training data can complement the task-specific information learned by the model during fine-tuning on a downstream task. In this way, we aim to help BERT not to forget assigning importance to informative input tokens when making predictions by proposing \textsc{SaLoss}; an auxiliary loss function for guiding the multi-head attention mechanism during training to be close to salient information extracted \textit{a priori} using TextRank. Experiments for explanation faithfulness across five datasets, show that models trained with \textsc{SaLoss} consistently provide more faithful explanations across four different feature attribution methods compared to vanilla BERT. Using the rationales extracted from vanilla BERT and \textsc{SaLoss} models to train inherently faithful classifiers, we further show that the latter result in higher predictive performance in downstream tasks.\footnote{Code: \url{https://github.com/GChrysostomou/saloss}.}
\end{abstract}

\section{Introduction}

Pretrained transformer-based \citep{NIPS2017_7181} language models (LMs)  such as \textsc{Bert} \citep{devlin-etal-2019-bert}, have achieved state-of-the-art results in various language understanding tasks \citep{wang2018glue,wang2019superglue}. Despite their success, their highly complex nature consisting of millions of parameters, makes them difficult to interpret \citep{jain2020learning}. This has motivated new research on understanding and explaining their predictions. 

Previous work has explored whether LMs encode syntactic knowledge by studying their multi-head attention distributions \citep{clark2019does, htut2019attention, voita-etal-2019-analyzing}. Recent studies have evaluated the faithfulness of explanations\footnote{A faithful explanation represents the true reasons behind a model's prediction~\citep{jacovi-goldberg-2020-towards}.} for predictions made by these models \citep{vashishth2019attention, atanasova2020diagnostic, jain2020learning}. In general, LMs can provide faithful explanations, particularly using attention~\citep{jain2020learning}, but still fall behind other simpler architectures~\citep{atanasova2020diagnostic} possibly due to increased information mixing and higher contextualization in the model~\citep{brunner2019identifiability, pascual2020telling, tutek-snajder-2020-staying}. 
Recent studies have attempted to improve the explainability of non transformer-based models, by guiding them through an auxiliary objective towards informative input importance distributions (e.g. human or adversarial priors) \citep{ross-et-al-2017, liu-avci-2019-incorporating, moradi-etal-2021-measuring}.


In a similar direction, we propose \textbf{Sa}lient \textbf{Loss} (\textsc{SaLoss}), an auxiliary objective that allows the multi-head attention of the model to learn from salient information (i.e. token importance) during training to reduce the effects of information mixing~\citep{pascual2020telling}. We compute a priori token importance scores \citep{xu-etal-2020-self} using \textsc{TextRank} \citep{mihalcea-tarau-2004-textrank} (i.e. an unsupervised graph-based method) and penalize the model when the attention distribution deviates from the salience distribution. Our contributions are as follows:

\begin{itemize}
     \item We demonstrate that models trained with \textsc{SaLoss} generate more faithful explanations in an input erasure evaluation.
    \item We finally show that rationales extracted from \textsc{SaLoss} models result in higher predictive performance in downstream tasks when used as the only input for training inherently faithful classifiers.
\end{itemize}

\section{Related Work}

\paragraph{Model Explainability}

Explanations can be obtained by computing importance scores for input tokens to identify which parts of the input contributed the most towards a model's prediction (i.e. feature attribution). A common approach to attributing input importance is by measuring differences in a model's prediction between keeping and omitting an input token \citep{robnik2008explaining, li2016understanding, nguyen2018comparing}. Input importance can also be obtained by calculating the gradients of a prediction with respect to the input~\citep{kindermans2016investigating,li-etal-2016-visualizing,integrated_gradients, bastings-filippova-2020-elephant}. We can also use sparse linear meta-models that are easier to interpret~\citep{ribeiro2016model,Lundberg2017}. Finally, recent studies propose using feature attribution to extract a fraction of the input as a rationale and then use it to train a classifier~\citep{jain2020learning, treviso-martins-2020-explanation}.

\paragraph{Faithfulness of Pretrained LM Explanations}

\citet{brunner2019identifiability} criticize the ability of attention in providing faithful explanations for the inner workings of a LM, by showing that constructed adversary attention maps do not impact significantly the predictive performance. \citet{pruthi2019learning} show similar outcomes by manipulating attention to attend to uninformative tokens. \citet{pascual2020telling} and \citet{brunner2019identifiability} argue that this might be due to significant information mixing in higher layers of the model, with recent studies showing improvements in the faithfulness of attention-based explanations by addressing this \citep{chrysostomou-aletras-2021-improving, tutek-snajder-2020-staying}.

\citet{atanasova2020diagnostic} evaluate faithfulness of explanations \citep{jacovi-goldberg-2020-towards} by removing important tokens and observing differences in prediction, showing that generally gradient-based approaches for transformers produce more faithful explanations compared to sparse meta-models \citep{ribeiro2016model}. 
However, transformer-based explanations are less faithful compared to simpler models due to their highly parameterized architecture. \citet{atanasova2020diagnostic} also show that explanation faithfulness does not correlate with how plausible it is (understandable by humans) corroborating arguments made by \citet{jacovi-goldberg-2020-towards}. \citet{jain2020learning} show that attention-based feature attributions, in general, outperform gradient-based ones. 

A different branch of studies introduced adversarial auxiliary objectives to influence attention-based explanations during training \citep{kennedy-etal-2020-contextualizing, wiegreffe2019attention, ijcai2017-371, liu-avci-2019-incorporating}. These objectives have typically been used as a tool for evaluating explanation faithfulness generated by attention \citep{kennedy-etal-2020-contextualizing, wiegreffe2019attention, pruthi2019learning, Ghorbani_Abid_Zou_2019} while others used auxiliary objectives to improve the faithfulness of explanations generated by non-transformer based models \citep{ijcai2017-371, liu-avci-2019-incorporating, moradi-etal-2021-measuring, mohankumar-etal-2020-towards, tutek-snajder-2020-staying}. The auxiliary objectives guide the model using human annotated importance scores \citep{liu-avci-2019-incorporating}, or allow for selective input gradient penalization \citep{ijcai2017-371}. Such studies illustrate the effectiveness of auxiliary objectives for improving the faithfulness of model explanations suggesting that we can also improve explanation faithfulness in transformers using appropriate prior information.

\section{Improving Explanation Faithfulness with Word Salience}

Even though attention scores are more faithful than other feature attribution approaches~\citep{jain2020learning}, they usually pertain to their corresponding input tokens in \textit{context} and not individually due to information mixing~\citep{tutek-snajder-2020-staying, pascual2020telling}. As such, we hypothesize that we can improve the ability of a pretrained LM in providing \emph{faithful} explanations,  by showing to the model alternative distributions of input importance (i.e. word salience). We assume that by introducing the salience distribution via an auxiliary objective \citep{ijcai2017-371}, we can reduce information mixing by ``shifting'' the model's attention to other informative tokens.
In a similar direction to ours, \citet{xu-etal-2020-self} showed that by computing attention together with salience information from keyword extractors improves text summarization.

\paragraph{Computing Word Salience}
We compute word salience $\boldsymbol \sigma$ using \textsc{TextRank} \citep{mihalcea-tarau-2004-textrank}, an unsupervised graph-based model for keyword extraction. \textsc{TextRank} calculates indegree centrality of graph nodes iteratively based on a Markov chain, where each node is a wordpiece and each edge links wordpiece pairs within a context window \citep{xu-etal-2020-self}. For each input document $X$, we construct an undirected graph and apply \textsc{TextRank} to compute the local salience scores ($\sigma_i$) of its words by:
\begin{equation}
    \sigma_i = (1-d) + d \displaystyle \sum_{j \in In(V_i)} \frac{\sigma_j}{|Out(V_j)|}
\end{equation}
\noindent where $d$ is the damping coefficient, $In(V_i)$ and $Out(V_j)$ are the incoming and outgoing nodes. Our intuition is that by using the task-agnostic \textsc{TextRank}, we can extract words that are important in the context of the sequence and as such offer an alternative view of token importance.\footnote{We also considered the use of \textsc{Tfidf} and \textsc{$\chi^2$} scores observing comparable but lower performance in early experimentation. We hypothesize that TextRank performs well due to its effectiveness in improving performance in text summarization \citep{xu-etal-2020-self}. See also Appx. \ref{appendix:input_erasure} for \textsc{Tfidf} and \textsc{$\chi^2$} results on input erasure experiments.} 

\paragraph{Salience Loss}

We propose Salient Loss (\textsc{SaLoss}), an auxiliary objective which allows the model to learn attending to more informative input tokens jointly with the task. \textsc{SaLoss} penalizes the model when the attention distribution ($\boldsymbol \alpha$) deviates from the word salience distribution ($\boldsymbol \sigma$).\footnote{$\boldsymbol \alpha \in \rm \!R^{t}$; $\boldsymbol \sigma \in \rm \!R^{t}$, where $t$ is the sequence length.} For $\boldsymbol \alpha$ we compute the average attention scores of the \texttt{CLS} token from the last layer \citep{jain2020learning}.
The joint objective for adapting a LM to a downstream classification task with \textsc{SaLoss} is:
\noindent 
\begin{equation}
    \mathcal{L} = \mathcal{L}_c  + \lambda \mathcal{L}_{sal}
\end{equation}
\noindent where $\mathcal{L}_c$ is the Cross-Entropy Loss for a downstream text classification task and $\lambda$ a regularization coefficient for the proposed \textsc{SaLoss} ($\mathcal{L}_{sal}$) which can be tuned in a development set.  $\mathcal{L}_{sal}$ is defined as the KL divergence between $\boldsymbol \alpha$ and $\boldsymbol \sigma$:
\noindent 
\begin{equation}
    \mathcal{L}_{sal} = KL(\boldsymbol \alpha, \boldsymbol \sigma) = \sum \boldsymbol \alpha (\log \boldsymbol \alpha - \log \boldsymbol \sigma)
\end{equation}

We assume a standard text classification setting where a set of labeled documents is used for fine-tuning a pretrained LM by adding an extra output classification layer. We normalize the salience scores for compatibility with the KL divergence.

\section{Experimental Setup}

\paragraph{Datasets}

We consider five natural language understanding tasks (see dataset statistics in Appx. \ref{appendix:datasets}): 
\textsc{SST} \citep{socher-etal-2013-recursive};  AGNews (\textsc{AG}) \citep{del2005ranking}; Evidence Inference (\textsc{Ev.Inf.}) \citep{lehman-etal-2019-inferring}; MultiRC (\textsc{M.RC}) \citep{khashabi-etal-2018-looking} and Semeval 2017 Task 4 Subtask A (\textsc{Semeval}) \citep{rosenthal-etal-2017-semeval}.


\paragraph{Models}

Similar to \citet{jain2020learning} we use: \textsc{BERT} \citep{devlin-etal-2019-bert} for (\textsc{SST}, \textsc{AG}, \textsc{SEMEVAL}); \textsc{SciBERT} \citep{beltagy-etal-2019-scibert} for \textsc{Ev.Inf.}; \textsc{Roberta} \citep{roberta_paper} for \textsc{M.RC}.

\paragraph{Evaluating Explanation Faithfulness}
\label{sec:interpretability}

We evaluate the faithfulness\footnote{We do not conduct human experiments, as faithfulness and plausibility (human understandability of explanations) do not correlate \citep{atanasova2020diagnostic, jacovi-goldberg-2020-towards, wiegreffe2019attention}.} of model explanations using two standard approaches: 
\begin{itemize}
    \item {\bf Input Erasure:}  We first compute the average fraction of tokens required to be removed (in decreasing importance) to cause a change in prediction (decision flip) \citep{serrano2019attention, nguyen-2018-comparing}.

    \item {\bf FRESH:} We also compute the predictive performance of a classifier trained on rationales extracted with feature attribution metrics (see $\S$\ref{ssec:feat_att}) using \textsc{FRESH} \citep{jain2020learning}. We extract rationales by; (1) selecting the top-k most important tokens (\textsc{TopK}) and (2) selecting the span of length \textit{k} that has the highest overall importance (\textsc{Contiguous}). 
\end{itemize}

\floatsetup[table]{font=sc}
\begin{table}[!t]
\small
\centering
\begin{tabular}{l||ccc}
 \textbf{Dataset} & Baseline & $\lambda$ & \textsc{SaLoss} \\ \hline \hline
 \textbf{SST} & .91 (.00) & 1e-3 & .91 (.00) \\
 \textbf{AG} & .93 (.00) & 1e-4 & .93 (.00) \\
 \textbf{Ev.Inf} & .82 (.01) & 1e-4 & .80 (.02) \\
 \textbf{M.RC} &  .76 (.01)  & 1e-3 & .76 (.00) \\
 \textbf{SEMEVAL} & .58 (.01)  & 1e-3 & .57 (.03)  
\end{tabular}
\caption{F1 macro averaged across 3 seeds for vanilla LMs (\textsc{Baseline}) and \textsc{SaLoss} models. $\lambda$ represents the regularization coefficient of our proposed objective.}
\label{tab:predictive_performances}
\end{table}

\paragraph{Feature Attribution Approaches} 
\label{ssec:feat_att}
We opt using the following popular metrics to allocate importance to input tokens: (1) Normalized attention scores ($\boldsymbol \alpha$); (2) Attention scores scaled by their gradient ($\boldsymbol \alpha \nabla \boldsymbol \alpha$) \citep{serrano2019attention}; (3) Gradients of the input scaled by the input ($\mathbf{x} \nabla \mathbf{x}$) \citep{kindermans2016investigating,atanasova2020diagnostic}; and (4) Integrated Gradients which compute the accumulated gradients along a path from a baseline to the input (\textbf{I.G.}) \citep{integrated_gradients}.\footnote{where $\nabla \alpha_i  = \frac{\partial \hat{y}}{\partial \alpha_i}$ and $\nabla x_i  = \frac{\partial \hat{y}}{\partial x_i}$}


\section{Experimental Results}

\paragraph{Predictive Performance}

\floatsetup[table]{font=sc}
\renewcommand*{\arraystretch}{1.00}
\begin{table}[!t]
\small
\centering
\setlength{\tabcolsep}{2pt}

\begin{tabular}{cc||ccccc}
       & \textbf{Metric} & \textbf{SST}  & \textbf{AG}  & \textbf{Ev.Inf.}  & \textbf{M.Rc} & \textbf{SEMEVAL} \\ \hline \hline
    & \textbf{Rand.} & .66 & .67 & .51 & .44 & .54 \\ \hline
     \parbox[t]{2mm}{\multirow{4}{*}{\rotatebox[origin=c]{90}{BaseLine}}} & $\alpha$ & .55 &  .43 & .25 & .40 & .43 \\
  &  $\mathbf{x} \nabla \mathbf{x}$ & .65 & .64  & .42 & .40 & .55 \\
    & $\alpha \nabla \alpha$ & .57 & .52  & .25 & .38 & .48  \\
    & \textbf{I.G.} & .63 &  .63 & .42 & .42 & .50  \\ \hline 
    \parbox[t]{2mm}{\multirow{4}{*}{\rotatebox[origin=c]{90}{SaLoss}}} & $\alpha$ &  \textbf{.42}$\dagger$ & .53  & .14$\dagger$ & \textbf{.19}$\dagger$ &  \textbf{.39}$\dagger$   \\
  &  $\mathbf{x} \nabla \mathbf{x}$ & .61$\dagger$ & .59$\dagger$  & .38$\dagger$ & .30$\dagger$ &  .51$\dagger$  \\
    & $\alpha \nabla \alpha$ & .48$\dagger$ & .50$\dagger$  & \textbf{.12}$\dagger$ & .24$\dagger$ & .41$\dagger$   \\
     & \textbf{I.G.}  & .61$\dagger$ &  .57$\dagger$ & .33$\dagger$ & .33$\dagger$ &  .45$\dagger$ 
    
\end{tabular}
\caption{Average fraction of tokens required to cause a decision flip across datasets and feature attribution metrics (lower is better). \textbf{Bold} denotes the best method in each dataset. $\dagger$ denotes a significant difference compared to \textsc{BASELINE} using the same attribution metric (Wilcoxon Rank Sum, $p < .05$).}
\label{tab:erasure_results}
\end{table}

Table~\ref{tab:predictive_performances} shows F1 macro scores averaged over three runs with standard deviation across tasks, for vanilla pretrained LMs (\textsc{Baseline}) and models with our proposed objective \textsc{SaLoss}. Results demonstrate that models trained with our proposed salience objective\footnote{We treat $\lambda$ as a hyper-parameter tuned on the development set, where $\lambda \in$ \{1e-2, 1e-3, 1e-4\}.} achieve similar performance to the \textsc{Baseline} models across datasets.

\paragraph{Input Erasure}

Table \ref{tab:erasure_results} shows results for the average fraction of input tokens required to be removed to cause a decision flip for  \textsc{Baseline} and \textsc{SaLoss} models in the test set. 
Results suggest that models trained with our proposed objective require a significantly lower fraction of tokens removed to cause a decision flip in 19 out of 20 cases (Wilcoxon Rank Sum, $p < .05$), with the exception of \textsc{AG} and $\alpha$.
This demonstrates that \textsc{SaLoss} obtains more faithful explanations in the majority of cases~\cite{jacovi-goldberg-2020-towards}. 
For example in \textsc{Ev.Inf.}, the \textsc{Baseline} approach with $\alpha$ requires  .25 fractions of tokens on average to observe a decision flip compared to .14 with \textsc{SaLoss} (approximately 40 tokens less).
We also observe that in \textsc{M.Rc.} where $\boldsymbol \alpha$ is not the most effective feature attribution method with \textsc{Baseline}, with \textsc{SaLoss} it becomes the most effective. In fact, $\boldsymbol \alpha$ is the best performing feature attribution approach across most tasks and metrics using \textsc{SaLoss}, indicating the effectiveness of infusing salient information. 

We also performed an analysis on the differences in Part-of-Speech (PoS) tags of the rationales selected by \textsc{SaLoss} and \textsc{Baseline},to obtain insights towards why rationales with \textsc{SaLoss} are shown to be more faithful to those from models trained without our proposed objective . In \textsc{SST}, we observe that \textsc{SaLoss} allocates more importance on adverbs and adjectives, which are considered important in sentiment analysis~\citep{dragut-fellbaum-2014-role, sharma-etal-2015-adjective}.
In \textsc{Ev.Inf.}, we observe that \textsc{SaLoss} allocates importance to subordinating conjunction  words such as \emph{than}, which are indeed important for the task, which consists of inferring relationships (i.e. \emph{higher than}). 
We thus hypothesize that \textsc{SaLoss} guides the model to other informative tokens, complementing the task specific information learned by the model.\footnote{We include an extensive analysis in Appx. \ref{appendix:pos}.}

\paragraph{Rationale Extraction}

We finally compare our \textsc{SaLoss} models with vanilla LMs (\textsc{BASELINE}) on rationale extraction using \textsc{FRESH} \citep{jain2020learning}, by measuring the predictive performance of the classifier trained on the extracted rationales. For completeness we also include an uninformative baseline for \textsc{SaLoss}, which comprise of a normalized uniform distribution over the input (i.e. all inputs are assigned the same salience score). For brevity, Table~\ref{tab:pred_performance_sali} presents results using the best performing metric from the erasure experiments $\alpha$ with \textsc{TopK}.\footnote{For \textsc{Contiguous} see Appx. \ref{appendix:FRESH}} 
Our approach significantly outperforms \textsc{Baseline} in 2 out of 5 datasets (t-test, $p<0.05$), whilst achieving comparable predictive performance on the rest. For example in \textsc{SST} we observe a 3\% increase in F1 using the same ratio of rationales. 
It is notable that in  \textsc{M.Rc}, \textsc{AG} and \textsc{Ev.Inf.}, performance of classifiers trained on rationales from both \textsc{Base.} and \textsc{SaLoss} is comparable to that with full text (1-2\% lower). 
We assume that this is due to the nature of the tasks, which likely do not require a large part of the input to reach high performance. This highlights the effectiveness of our approach, as a simple yet effective solution for improving explanation faithfulness. 


\renewcommand*{\arraystretch}{1.0}
\begin{table}[!t]
\small
\centering
\setlength{\tabcolsep}{1.5pt}
\begin{tabular}{l||c|cc}
      \multirow{2}{*}{\textbf{Dataset}} & \multirow{2}{*}{\textbf{Baseline}} & \multicolumn{2}{c}{\textbf{SaLoss}}\\
        &  & \textbf{TextRank} & \textbf{Uniform} \\\hline \hline
      \textbf{SST} (20\%) &  .83 (.00)   & \textbf{.87} (.00) $\dagger$ & .82 (.00)\\
      \textbf{AG} (20\%) & .92 (.00)  & .92 (.00) & .92 (.00) \\
      \textbf{Ev.Inf.} (10\%) &  \textbf{.82} (.00) & .81 (.00) & .78 (.00) \\
      \textbf{M.Rc} (20\%) &  .75 (.00)  & .75 (.00) & .75 (.00)  \\
      \textbf{SEMEVAL} (20\%)  &  .48 (.03) & \textbf{.53} (.01)$\dagger$ & .43 (.00) \\
\end{tabular}
\caption{F1 macro on models trained with extracted rationales (\textsc{TopK} and $\alpha$) using FRESH for  \textsc{Baseline} and \textsc{SaLoss} models. \textbf{Bold} denotes best performance in each dataset. $\dagger$ indicates that \textsc{SaLoss} rationales perform significantly better (t-test, $p < .05$).}
\label{tab:pred_performance_sali}
\end{table}
\floatsetup[table]{font=sf}
\begin{table*}[!t]
    
    \scriptsize
    \centering
    \begin{tabular}{p{\linewidth}}
        \sffamily
        \\
        
         \multicolumn{1}{l}{\scriptsize\textbf{Example 1}} \textbf{Data.:\textsc{AG} Id: test\_239} \\
         
         \textbf{[\textsc{Baseline}]:}   NEW YORK ( Reuters ) - Shares of \hl{\scriptsize\textbf{Google Inc. will make their Nasdaq stock market}} debut on Thursday after the year 's most   anticipated initial public offering priced far below initial estimates , raising \$1.67 billion . \\
         
         \textbf{[\textsc{SaLoss} (Ours)]:}  NEW YORK ( Reuters ) - Shares of Google Inc. will make their   Nasdaq stock market debut on Thursday after the year 's most   anticipated initial public offering \hl{\scriptsize\textbf{priced far below initial estimates , raising \$1.67}} billion . \\
         
         \textbf{[Topic]:} Business \\ \hline \hline

        
        \multicolumn{1}{l}{\scriptsize\textbf{Example 2}} \textbf{Data.:\textsc{SST} Id: test\_78}\\

        \textbf{[\textsc{Baseline}]:} If nothing else this \hl{\scriptsize\textbf{movie introduces a}} promising unusual kind of psychological horror.
        
        \textbf{[\textsc{SaLoss} (Ours)]:} If nothing else this movie introduces \hl{\scriptsize\textbf{a promising unusual}} kind of psychological horror.
        
        \textbf{[Sentiment]:} Positive  \\ \hline \hline
         
         \multicolumn{1}{l}{\scriptsize\textbf{Example 3}} \textbf{Data.:\textsc{Ev.Inf.} Id: 4118506\_0}\\
        \textbf{[\textsc{Baseline}]:} ... analgesics . \hl{\scriptsize\textbf{ABSTRACT.AIM : : The aim of this study is to evaluate the efficacy of fentanyl along with LA field infiltration in controlling pain and discomfort associated with CVC insertion . ABSTRACT}}.SETTINGS AND DESIGN : :...
        
        \textbf{[\textsc{SaLoss} (Ours)]:} ... ABSTRACT.RESULTS : : The median interquartile range pain score is worst for placebo group after LAI ( 5 [ 3 - 6 ] ) and in the immediate postprocedure period ( 5 [ 4 - 5 ] ) \hl{\scriptsize\textbf{which was significantly attenuated by addition of fentanyl ( 3.5 [ 2 - 5 ] and 3 [ 2 - 4 ] ) ( P = 0.009 and 0.001}} respectively ) ...\\
       
         \textbf{[Intervention || Comparator || Outcome]:} Fentanyl \textbf{||} Normal saline \textbf{||} Pain score \\ 
         
         \textbf{[Relationship]:} Significantly decreased  \\ \hline \hline

    \end{tabular}

    \caption{True examples of extracted rationales from models using our proposed approach (\textsc{SaLoss}) and from models that do not (\textsc{Baseline})}
    \label{fig:examples}
    
\end{table*}

\section{Qualitative Analysis}

In Table \ref{fig:examples} we present examples of extracted rationales from a model trained with our proposed objective (\textsc{SaLoss}) and without (\textsc{Baseline}) using $\alpha\nabla\alpha$, to gain further insights to complement the PoS analysis. For clarity we present rationales of \textsc{Contiguous} type. 

In AG we observed similar performance between models trained with \textsc{SaLoss} and without. Example 1 illustrates such a case, where both models predicted correctly but attended to different parts of the input. Despite in different locations, both segments are closely associated with the label of ``Business''.
Example 2 is an instance from the SST dataset, were the \textsc{SaLoss} rationale points to a phrase that is more associated with the task (``\textit{a promising unusual}'') compared to the \textsc{Baseline}. This also aligns with previous observations from the PoS analysis, that models trained with our proposed objective attend to more adjectives compared to \textsc{Baseline}. 
Example 3 considers an instance from the Ev.Inf. dataset, which shows that the model trained with \textsc{SaLoss} and \textsc{Baseline} attended to two different sections. In fact what we observed in agreement with the PoS analysis, is that models with \textsc{SaLoss} attend mostly to segments including words related to relationships, such as ``\textit{significantly attenuated}'' in this particular example. 

\section{Conclusion}

We introduced \textbf{Sa}lient \textbf{Loss} (\textsc{SaLoss}), an auxiliary objective to incorporate salient information to attention for improving the faithfulness of transformer-based prediction explanations. We demonstrate that our approach provides more faithful explanations compared to vanilla LMs on input erasure and rationale extraction. In the future, we plan to explore additional objectives to better optimize for contiguity of rationales.

\section*{Acknowledgments}
NA is supported by EPSRC grant EP/V055712/1, part of the European Commission CHIST-ERA programme, call 2019 XAI: Explainable Machine Learning-based Artificial Intelligence.

\bibliographystyle{acl_natbib}
\bibliography{biblio}

\clearpage
\newpage
\begin{appendices}

\section{Datasets \label{appendix:datasets}}

For our experiments we use the following tasks (see dataset details in Table \ref{tab:data_characteristics}.):

\paragraph{\textbf{SST}} \citep{socher-etal-2013-recursive}: Binary sentiment classification with removed neutral sentences.
\paragraph{\textbf{AG}} News \citep{del2005ranking}: News articles categorized by the following topics; Science, Sports, Business, and World.
\paragraph{\textbf{Ev.Inf}} (Evidence Inference) \citep{lehman-etal-2019-inferring}:  Abstract-only biomedical  articles  describing  randomized  controlled trials.  The task is to infer the reported relationship between a given intervention and comparator with respect to an outcome.
\paragraph{\textsc{M.Rc}} (Multi RC) \citep{khashabi-etal-2018-looking}:  A reading comprehension  dataset  composed  of  questions  with multiple correct answers that depend on information from multiple sentences. Similar to \citet{deyoung-etal-2020-eraser} and \citet{jain2020learning} we convert this to a binary classification task where  each rationale/question/answer triplet forms an instance and each candidate answer has a label of True or False.
\paragraph{SEMEVAL} \citet{rosenthal-etal-2017-semeval}: The Semeval 2017 dataset for Task 4 Subtask A which consists of tweets and the task is to classify whether the message is of positive, negative, or neutral sentiment.

\floatsetup[table]{font=sc}
\renewcommand*{\arraystretch}{1.0}
\begin{table}[!b]
\setlength\tabcolsep{3pt}
\small
\centering
\begin{tabular}{l||ccc}
\textbf{Data}   & Av. $|W|$ & \textbf{C} & \textbf{\begin{tabular}[c]{@{}c@{}}Splits\\   Train/Dev/Test\end{tabular}} \\ \hline
\textbf{SST}                & 18                  & 2     & 6,920 / 872 / 1,821  
\\
\textbf{AG}         & 36                  & 4   & 102,000 / 18,000 / 7,600         \\
\textbf{Ev.Inf.}               & 363                 & 3     & 5,789 / 684 / 720  \\
\textbf{M.RC}             & 305                  & 2     & 24,029 / 3,214 / 4,848   \\

\textbf{SEMEVAL} & 20 & 3 &  6,000 / 2,000 / 20,630  
\end{tabular}
\caption{Dataset statistics including average words per input, number of classes and splits (see also Appx. \ref{appendix:datasets}).}
\label{tab:data_characteristics}
\end{table}

\section{TextRank Training 
\label{appendix:textrank}}

We run for 10 steps, or until convergence, with a window of 4 words, a damping coefficient of 0.85 and normalize the salience scores to make them more compatible to attention distributions.

\section{Model Hyper-Parameters \label{appendix:model_parameters}}

Table \ref{tab:model_hyperparameters} presents the hyper-parameters used to train the models across different datasets, along with F1 macro performance on the development set. Models where finetuned across 3 runs for 10 epochs, with the exception of the \textsc{SEMEVAL} dataset which was finetuned for 20. We implement our models using the Huggingface library \citep{Wolf2019HuggingFacesTS} and use default parameters of the \textsc{AdamW} optimizer apart from the learning rates and a linear scheduler. Each experiment is run on a single Nvidia Tesla V100 GPU.

We found that the learning rate of our proposed objective, does not impact significantly F1 macro performance. As such, since our objective is improving faithfulness, our $\lambda$ selection includes training then evaluating on the development set the average fraction of tokens required to cause a decision flip. We use the model with the lowest fraction of tokens scores and report on the test set. 

\floatsetup[table]{font=sc}
\renewcommand*{\arraystretch}{1.2}
\begin{table}[!t]
\setlength\tabcolsep{2.5pt}
\small
\centering
\begin{tabular}{l||ccc|c}
\textbf{Dataset}   & Model & $lr^m$ & $lr^c$ &  F1 \\ \hline
SST                & bert-base & 1e-5 & 1e-4     &  .91  $\pm$ .00   \\
AG         & bert-base & 1e-5 & 1e-4     &  .93 $\pm$ .00      \\
Ev.Inf.             & scibert & 5e-6 & 2e-4   & .84 $\pm$ .01       \\
M.RC           & roberta-base & 2e-6 & 2e-4    &  .75 $\pm$ .01      \\

SEMEVAL & bert-base & 1e-5 & 1e-4    &  .59 $\pm$ .02      \\

\end{tabular}
\caption{Model and their hyper-parameters for each dataset, including learning rate for the model ($lr^m$) and the classifier layer ($lr^c$)  and F1 macro scores on the development set across three runs.}
\label{tab:model_hyperparameters}
\end{table}

\section{Further Details on Evaluating Faithfulness \label{appendix:faithfulness}}

\paragraph{\textbf{Erasure} \citep{serrano2019attention, nguyen-2018-comparing}:} \citet{jacovi-goldberg-2020-towards} propose that an appropriate measure of \emph{faithfulness} of an explanation can be obtained through \emph{input erasure} (the most relevant parts of the input--according to the explanation--are removed). We therefore record the average  fraction of tokens required to be removed across instances to cause a decision flip. Removal is conducted in descending token importance order at every 5\% of the length in the sequence, as searching at every token is computationally expensive \citep{atanasova2020diagnostic}. Note that we conduct all experiments at the input level (i.e. by removing the token from the input sequence instead of only removing its corresponding attention weight) as we consider the scores from importance metrics to pertain to the corresponding input token following related work \citep{larras2016explaining,arras2017explaining,nguyen2018comparing, vashishth2019attention,grimsley-etal-2020-attention}.

\paragraph{\textbf{FRESH} \citep{jain2020learning}:} A pipeline composed of a \textit{support model}-\textit{extractor}-\textit{classifier}, whereby the \textit{support model} is the model trained on the full text and allocates importance to tokens, \textit{extractor} the approach used and extract the rationales according to the importance from the \textit{support model} and \textit{classifier} the model trained on the rationales. The higher the \textit{classifier}'s predictive performance the more faithful the rationales by the \textit{support model}.

Similar to \citet{jain2020learning}, for \textsc{FRESH} we extract rationales  of a fixed ratio compared to the sequence length by two thresholder approaches (\textsc{Thresh.}):

\begin{itemize}
    \item \textsc{TopK}: The \emph{top-k} tokens as indicated by the corresponding importance metric, treating each word independently.
    \item \textsc{Contiguous}: The span of length \emph{k} that results in the highest overall score as indicated by the importance metric.
\end{itemize}

\section{Further Details on Feature Attribution Approaches \label{appendix:feature_attribution}}

\begin{itemize}
    \item $\boldsymbol \alpha$:  Importance rank corresponding to normalized attention scores. 

    \item $\boldsymbol \alpha \nabla \boldsymbol \alpha$: Scales the attention scores $\alpha_i$ with their corresponding gradients $\nabla \alpha_i  = \frac{\partial \hat{y}}{\partial \alpha_i}$. \citet{serrano2019attention}\footnote{\citet{serrano2019attention} show that gradient-based attention ranking metrics ($\boldsymbol \alpha \nabla \boldsymbol \alpha$) are better in providing faithful explanations compared to just using attention ($\boldsymbol \alpha$).} 
    
    \item $\mathbf{x} \nabla \mathbf{x}$ (InputXGrad) \citep{kindermans2016investigating,atanasova2020diagnostic}: Ranking words by multiplying the gradient of the input by the input with respect to the predicted class, where $\nabla x_i = \frac{\partial \hat{y}}{\partial x_i}$

    \item \textbf{$\mathbf{I.G.}$} (Integrated Gradients)  \citep{integrated_gradients}: Ranking words by computing the integral of the gradients taken along a straight path from a baseline input to the original input, where the baseline input is a sequence of zero embedding vectors.
    
\end{itemize}

\section{Further Results on FRESH \label{appendix:FRESH}}

In Table \ref{tab:FRESH_appendix} we present complementary results on the F1 macro scores of the classifier trained on extracted contiguous rationales. Rational ratios for datasets \textsc{SST, AG, Ev.Inf.} and \textsc{M.Rc} are from \citet{jain2020learning}, whilst for SEMEVAL we choose a 20\% ratio. 

We can first observe that models trained on contiguous rationale extracted from models trained with \textsc{SaLoss}, obtain comparable performance to models without (\textsc{Base}). Additionally, results show that classifier performance does not reach those with \textsc{TopK} rationales. We can therefore assume that \textsc{TopK} rationales result to inherently faithful classifiers with higher performance. It is encouraging to notice that in the datasets where performance is comparable with our approach (AG, \textsc{Ev.Inf.}, \textsc{M.Rc}), it is likely due to reaching close to \textsc{Full-Text} performance. For example, classifier performance trained on \textsc{Contiguous} rationales from \textsc{Base.} in SST is at .82 compared to .83 with \textsc{SaLoss} rationales. 

Results also suggest that our uninformative baseline (\textsc{Unif.}), reduces the faithfulness of rationales in most cases resulting in lower classifier performance. We hypothesize that in cases where performance is comparable with \textsc{Base.} and \textsc{SaLoss}, it is due to the task being relatively easy and as such the loss function not impacting the faithfulness of rationales. We consider this direction as an interesting area for future work.

\renewcommand*{\arraystretch}{1.1}
\begin{table}[!t]
\small
\centering
\setlength{\tabcolsep}{1.5pt}
\begin{tabular}{l||c|cc}
      \multirow{2}{*}{\textbf{Dataset}} & \multirow{2}{*}{\textbf{Baseline}} & \multicolumn{2}{c}{\textbf{SaLoss}}\\
        &  & \textbf{TextRank} & \textbf{Uniform} \\\hline \hline
      \textbf{SST} (20\%) &  .82 (.00)   & \textbf{.83} (.00) $\dagger$ & .80 (.00)\\
      \textbf{AG} (20\%) & \textbf{.90} (.00)  & .89 (.00) & .89 (.00) \\
      \textbf{Ev.Inf.} (10\%) &  \textbf{.79} (.00) & .78 (.00) & .78 (.00) \\
      \textbf{M.Rc} (20\%) &  .70 (.00)  & .67 (.00) & \textbf{.71} (.00)  \\
      \textbf{SEMEVAL} (20\%)  &  .46 (.03) & \textbf{.47} (.01)$\dagger$ & .42 (.00) \\
\end{tabular}
\caption{F1 macro on models trained with extracted rationales (\textsc{Contiguous} and $\alpha$) using FRESH for  \textsc{Baseline} and \textsc{SaLoss} models. \textbf{Bold} denotes best performance in each dataset. $\dagger$ indicates that \textsc{SaLoss} rationales perform significantly better (t-test, $p < 0.05$).}
\label{tab:FRESH_appendix}
\end{table}

\section{PoS Importance Allocation \label{appendix:pos}}

\begin{figure*}
     \centering
     \begin{subfigure}[b]{0.95\textwidth}
         \centering
         \includegraphics[width=\textwidth, height = 200pt]{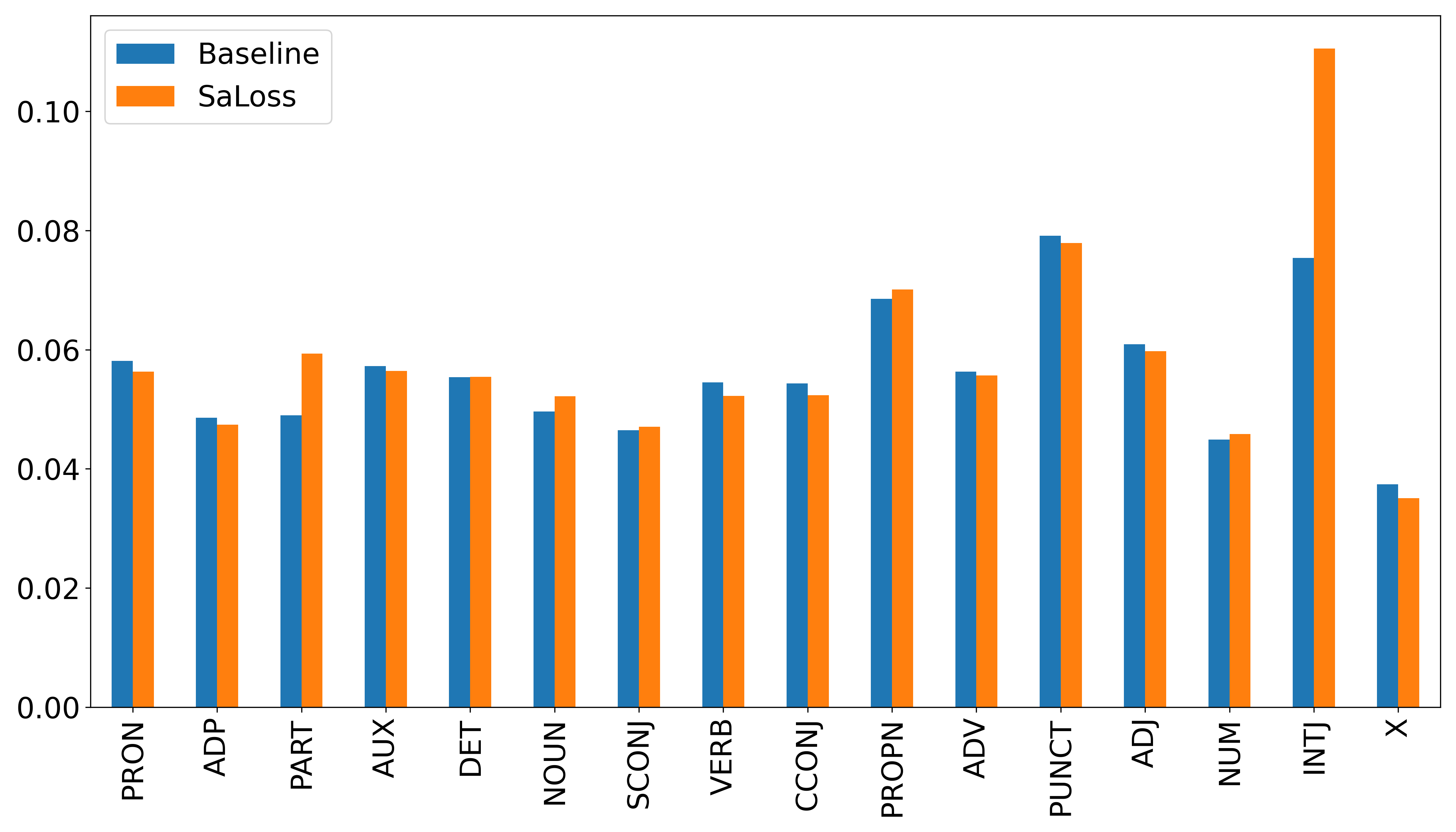}
         \caption{(\textsc{SST})}
         \label{subfig:sst}
     \end{subfigure}
     \vspace{0.5pt}
     \begin{subfigure}[b]{\textwidth}
         \centering
         \includegraphics[width=\textwidth, height = 200pt]{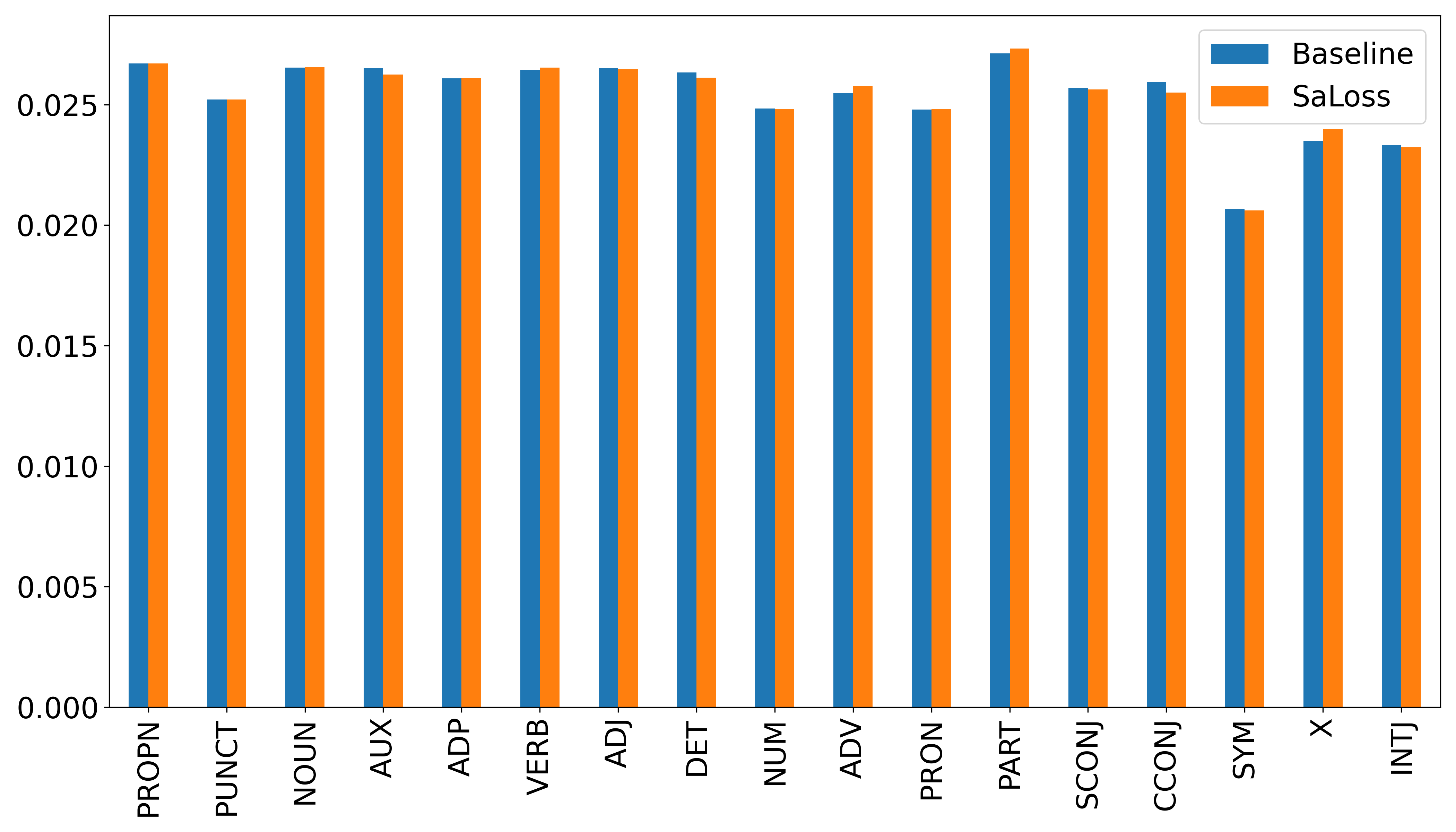}
         \caption{(\textsc{AG})}
         \label{subfig:agnews}
     \end{subfigure}
     \vspace{0.5pt}
     \begin{subfigure}[b]{\textwidth}
         \centering
         \includegraphics[width=\textwidth, height = 200pt]{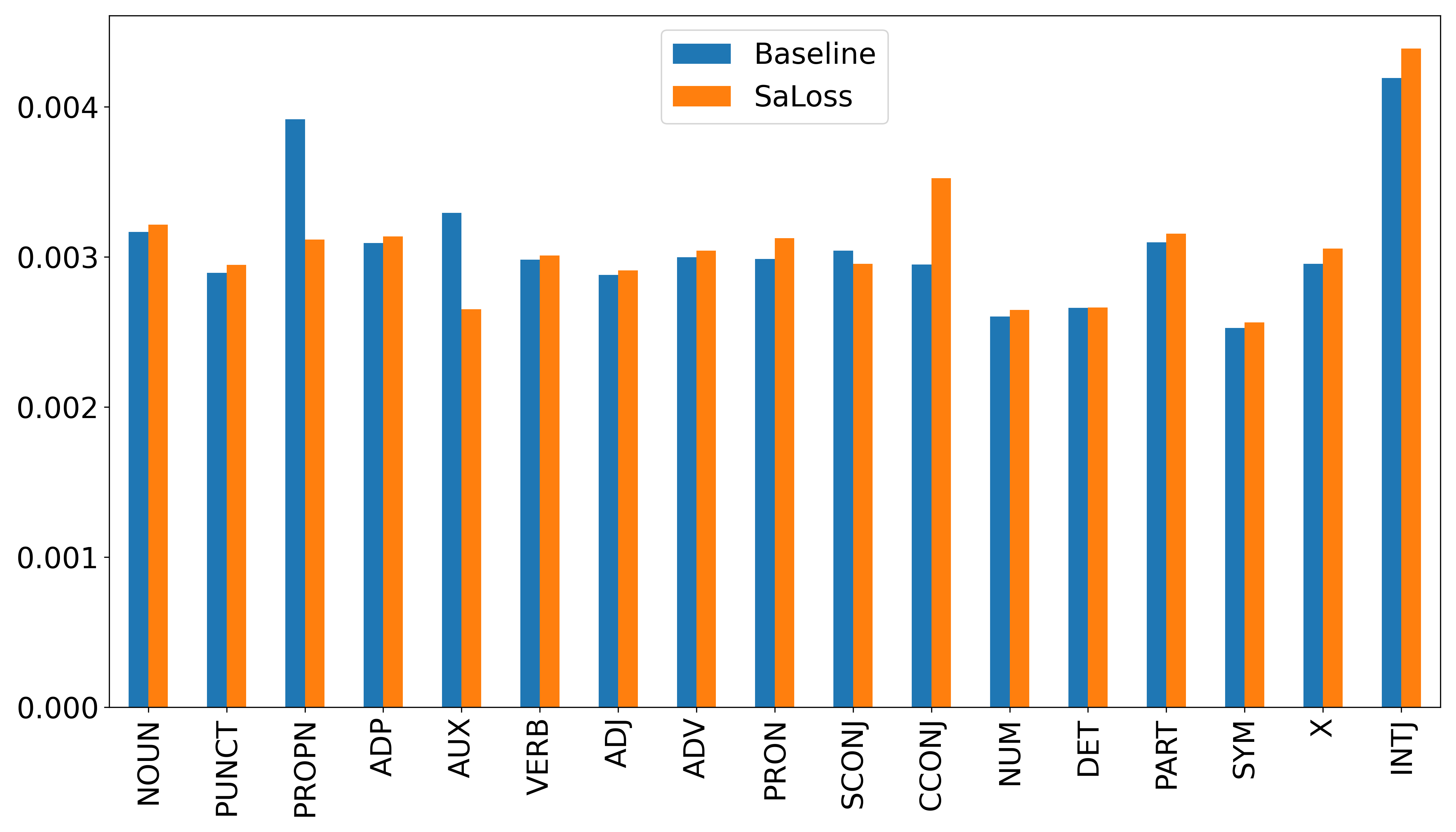}
         \caption{(\textsc{Ev.Inf.})}
         \label{subfig:evinf}
     \end{subfigure}
            \caption{Average importance across Part of Speech (PoS) tags as indicated by $\alpha\nabla\alpha$ with \textsc{Baseline}, with our proposed component \textsc{SaLoss}.}
        \label{fig:pos_tags}
\end{figure*}

We also conduct an analysis whereby we record the average importance scores under each Part of Speach (PoS) tag. We run a pretrained PoS tagger from \texttt{spaCy} \citep{spacy} across the text and compute average importance calculated from a feature attribution approach for each PoS tag. We therefore aim to observe differences in allocation of importance in linguistic features between models trained with out our proposed approach (\textsc{Base.}) and with (\textsc{SaLoss}). In Figure \ref{fig:pos_tags} we present distribution of importance (calculated with $\alpha\nabla\alpha$) across PoS tags, on three datasets (\textsc{SST}, \textsc{AG} and \textsc{Ev.Inf.}). 

Observing Figure \ref{subfig:sst}, we can see that $\alpha\nabla\alpha$ with \textsc{SaLoss} places greater importance on proper nouns (PROPN), auxiliary words (AUX), pronouns (PRON) and interjections (INTJ). In comparison the most prominent tags with \textsc{Base} are INTJ, PROPN, coordinating conjunctions (CCONJ) and nouns (NOUN). 
In a sentiment analysis task, it is notable that both \textsc{Base.} and \textsc{SaLoss} base high importance on average on interjections, which typically demonstrate feelings or emotions. Both appear to highlight particularly well adjectives, which we consider more important for sentiment analysis as they name attributes of other words. On the other end we also observe that \textsc{SaLoss} places lower importance on average to CCONJ and punctuation (PUNCT) compared to \textsc{Base.} This suggests that for \textsc{SST}, \textsc{SaLoss} models possibly shift their importance to more informative for the task word groups. 

Moving on to Figure \ref{subfig:agnews}, we observe a very high peak on proper nouns (PROPN) and unidentified tokens (X) with \textsc{SaLoss} compared to \textsc{Base.}. In a news classification task proper nouns such as the NATO and other organization or city names can indicate the topic of a sequence. We assume that for \textsc{SaLoss} to place such great importance on proper nouns, we manage with our approach to shift the model's attention to more informative for the task tokens. However we also observe unidentified symbols having large average importance scores with \textsc{SaLoss}. Whilst we do not study plausibility (human understandability of explanations), we consider this a limitation and we consider exploring and addressing this an interesting direction for future work. 

Finally, examining Figure \ref{subfig:evinf}, we observe that both \textsc{SaLoss} and \textsc{Base} place very high importance on particle (PART) words such as \textit{not}. We consider this encouraging, as large parts of the task is to infer if there was a significant difference or \emph{not} based on an observation in the text. Additionally, we observe that \textsc{SaLoss} attends highly to subordinating conjunction (SCONJ) words such as than, which if placed in the context of "\emph{significantly higher than}" directly relates to our task. Also with \textsc{SaLoss} we observe a reduction in attention to pronouns (PRON) compared to \textsc{Base}, which we consider encouraging as PRON words are not directly related to the task of infering relationships. This indicates that our proposed objective manages to guide the model's attention away from uninformative tokens such as others and punctation, and towards more informative for the task token types (SCONJ, CCONJ).

\section{Input Erasure 
\label{appendix:input_erasure}}

Table \ref{tab:erasure_results_appendix} presents the average fraction of tokens required to cause a prediction switch (decision flip), when training models with \textsc{SaLoss} and (1) \textsc{TextRank}; (2) \textsc{ChiSquared}; (3) \textsc{Tfidf}. We observe that when models are regularized with \textsc{TextRank} scores, the feature attribution approaches result in a lower average fraction of tokens to cause a prediction switch compared to the other two salience functions. We also observe that \textsc{Tfidf} is comparable with \textsc{TextRank} in most cases, outperforming \textsc{Chisquared}. We hypothesize that \textsc{Tfidf} performs poorer than \textsc{TextRank} is due to the way these two approaches compute their ``importance'' scores. The first computes them globally, whilst the latter locally (at instance-level) which we assume is more beneficial for explanation faithfulness. 

\floatsetup[table]{font=sc}
\renewcommand*{\arraystretch}{1.50}
\begin{table}[!t]
\small
\centering
\setlength{\tabcolsep}{2pt}

\begin{tabular}{cc||ccccc}
       & \textbf{Metric} & \textbf{SST}  & \textbf{AG}  & \textbf{Ev.Inf.}  & \textbf{M.Rc} & \textbf{SEMEVAL} \\ \hline \hline
    & \textbf{Rand.} & .66 & .67 & .51 & .44 & .54 \\ \hline
    \parbox[t]{2mm}{\multirow{4}{*}{\rotatebox[origin=c]{90}{TextRank}}} & $\alpha$ &  .42 & .53  & .14 & .19 &  .39   \\
  &  $\mathbf{x} \nabla \mathbf{x}$ & .61 & .59  & .38 & .30 &  .51  \\
    & $\alpha \nabla \alpha$ & .48 & .50  & .12 & .24 & .41   \\
     & \textbf{I.G.}  & .61 &  .57 & .33 & .33 &  .45 \\ \hline 
     
     \parbox[t]{2mm}{\multirow{4}{*}{\rotatebox[origin=c]{90}{ChiSquared}}} & $\alpha$ & .49 &    .67 &   .29 &     .38 &     .44 \\
  &  $\mathbf{x} \nabla \mathbf{x}$ & .60 &    .59 &   .47 &     .34 &     .54 \\
    & $\alpha \nabla \alpha$ &  .61 &    .71 &   .28 &     .33 &     .49 \\
     & \textbf{I.G.}  & .58 &    .56 &   .48 &     .38 &     .47 \\ \hline

     \parbox[t]{2mm}{\multirow{4}{*}{\rotatebox[origin=c]{90}{Tfidf}}} & $\alpha$ &  .47 &    .43 &   .20 &     .33 &     .48 \\
  &  $\mathbf{x} \nabla \mathbf{x}$ & .62 &    .57 &   .41 &     .36 &     .57 \\
    & $\alpha \nabla \alpha$ &   .50 &    .47 &   .20 &     .37 &     .58 \\
     & \textbf{I.G.}  & .58 &    .56 &   .40 &     .38 &     .53 \\
    
\end{tabular}
\caption{Average fraction of tokens required to cause a decision flip across datasets and feature attribution metrics (lower is better).}
\label{tab:erasure_results_appendix}
\end{table}

\end{appendices}
\end{document}